\documentclass[letterpaper]{article} 
\usepackage[]{aaai23}  
\usepackage{times}  
\usepackage{helvet}  
\usepackage{courier}  
\usepackage[hyphens]{url}  
\usepackage{graphicx} 
\usepackage{natbib}  
\usepackage{caption} 
\usepackage{hyperref}
\usepackage{algorithm}
\usepackage{algorithmic}

\usepackage{newfloat}
\usepackage{listings}
\DeclareCaptionStyle{ruled}{labelfont=normalfont,labelsep=colon,strut=off} 
\floatstyle{ruled}
\newfloat{listing}{tb}{lst}{}
\floatname{listing}{Listing}
\title{Memory-aided Contrastive Consensus Learning for Co-salient Object Detection}
\author{
    Peng Zheng\textsuperscript{\rm 1},
    Jie Qin$^{\rm 1*}$,
    Shuo Wang\textsuperscript{\rm 2},
    Tian-Zhu Xiang\textsuperscript{\rm 3},
    Huan Xiong\textsuperscript{\rm 4,5}\thanks{Corresponding authors.}
}
\affiliations{
    \textsuperscript{\rm 1} Nanjing University of Aeronautics and Astronautics, Nanjing, China \\
    \textsuperscript{\rm 2} ETH Zurich, Zurich, Switzerland \\
    \textsuperscript{\rm 3} Inception Institute of Artificial Intelligence, Abu Dhabi, UAE \\
    \textsuperscript{\rm 4} Harbin Institute of Technology, China \\
    \textsuperscript{\rm 5} Mohamed bin Zayed University of Artificial Intelligence, Abu Dhabi, UAE \\
    
    \{zhengpeng0108, qinjiebuaa, shawnwang.tech, tianzhu.xiang19, huan.xiong.math\}@gmail.com
}

\usepackage{bibentry}

\usepackage{makecell}
\usepackage[table]{xcolor}
\usepackage{multirow}
\usepackage{amsfonts}
\usepackage{overpic}

\def\eg{\emph{e.g}.} 
\def\ie{\emph{i.e}.} 
 
\def\etc{\emph{etc}.}

\def\ourmodel{MCCL}
\def\numModel{13}
\def\fps{150}

\def\impSCoSOD3k{$\sim$5.9\%}
\def\impSCoSal2015{$\sim$6.2\%}

\newcommand{\figref}[1]{Fig.~\ref{#1}}
\newcommand{\tabref}[1]{Tab.~\ref{#1}}

\definecolor{mygray}{gray}{.92}

\makeatletter
\newcommand{\thickhline}{%
    \noalign {\ifnum 0=`}\fi \hrule height 1pt
    \futurelet \reserved@a \@xhline
}

\begin{document}

\maketitle

\begin{abstract}

Co-Salient Object Detection (CoSOD) aims at detecting common salient objects within a group of relevant source images. Most of the latest works employ the attention mechanism for finding common objects.
To achieve accurate CoSOD results with high-quality maps and high efficiency, we propose a novel Memory-aided Contrastive Consensus Learning (\ourmodel{}) framework, which is capable of effectively detecting co-salient objects in real time ($\sim$\fps{} fps).
To learn better group consensus, we propose the Group Consensus Aggregation Module (GCAM) to abstract the common features of each image group;
meanwhile, to make the consensus representation more discriminative, we introduce the Memory-based Contrastive Module (MCM), which saves and updates the consensus of images from different groups in a queue of memories.
Finally, to improve the quality and integrity of the predicted maps, we develop an Adversarial Integrity Learning (AIL) strategy to make the segmented regions more likely composed of complete objects with less surrounding noise.
Extensive experiments on all the latest CoSOD benchmarks demonstrate that our lite \ourmodel{}~outperforms \numModel{} cutting-edge models, achieving the new state of the art (\impSCoSOD3k{} and \impSCoSal2015{} improvement in S-measure on CoSOD3k and CoSal2015, respectively). Our source codes, saliency maps, and online demos are publicly available at https://github.com/ZhengPeng7/MCCL.
\end{abstract}

\section{Introduction}
\label{sec:introduction}

\begin{figure}[t!]
  \centering
    \begin{overpic}[width=.8\linewidth]{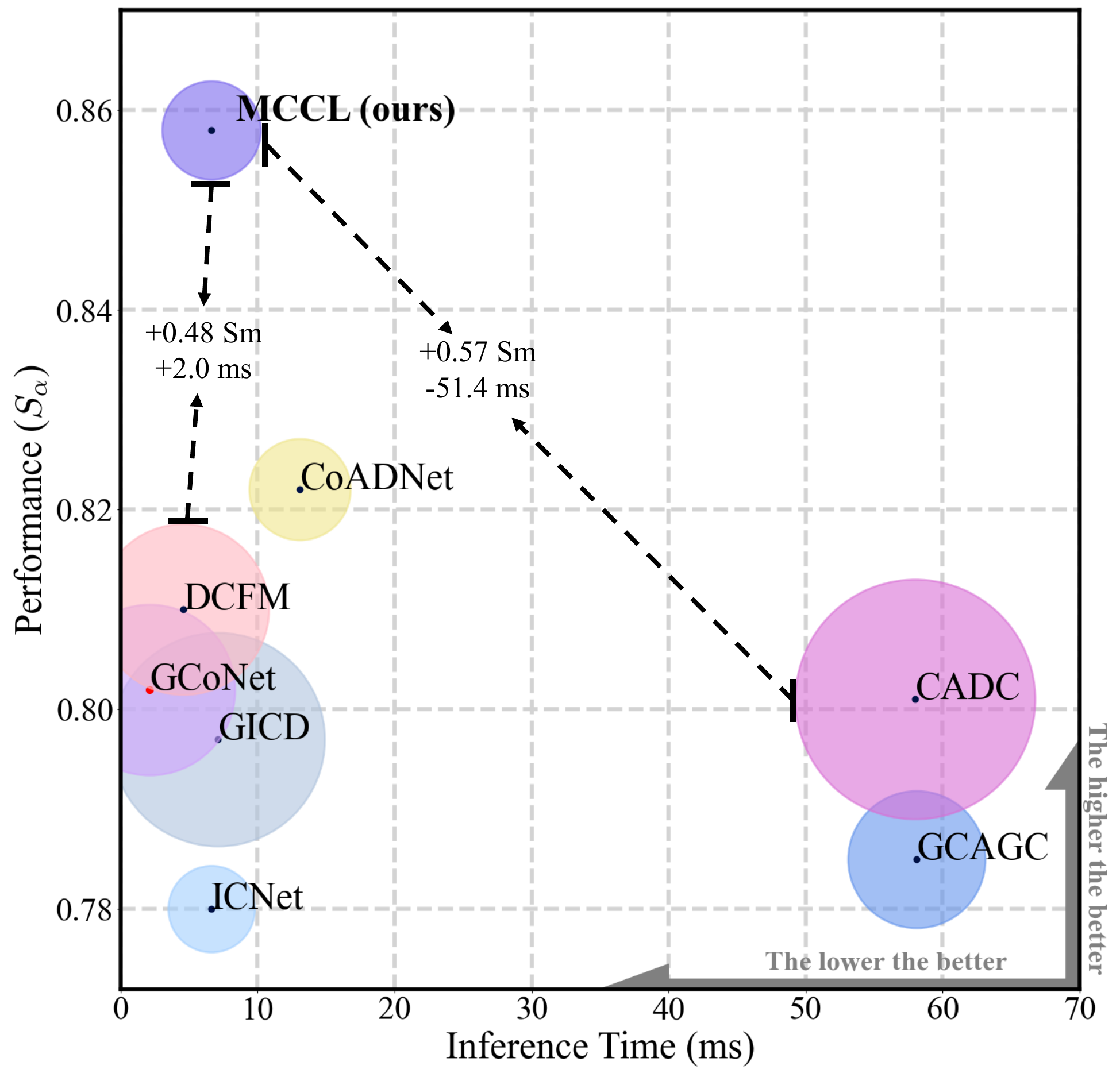}
    \end{overpic}
   \caption{
   \textbf{Accuracy (S-measure) and inference time (ms) of the update-to-date and representative deep-learning-based CoSOD methods on the CoSOD3k dataset~\cite{CoSOD3k}.} We conduct the comparison on both accuracy (the vertical axis) and speed (the horizontal axis) among seven existing representative CoSOD models and our \ourmodel{}. Larger bubbles mean larger volume of the model weights. Our \ourmodel{}~achieves great performance (0.860 S-measure) in real time (7.6 ms) with a lightweight model (104.5 Mb weights and 5.93G FLOPs). All the methods are tested with batch size 2 on one A100-80G, an online benchmark for inference speed can be found at \url{https://github.com/ZhengPeng7/CoSOD_fps_collection}.
   }\label{fig:acc_runtime}
\end{figure}

Co-Salient object detection (CoSOD) aims at detecting the most common salient objects among a group of source images. Compared with the standard salient object detection (SOD) task, CoSOD is more challenging for distinguishing co-occurring objects across a group of images where salient objects of different classes and non-salient objects of the same class are both hard distractors.
CoSOD methods also show their advantage of acting as a pre-processing step for other computer vision tasks, such as semantic segmentation~\cite{zeng2019joint}, 
co-segmentation~\cite{hsu2019deepco3}, 
and object tracking~\cite{Zhou2021SaliencyAssociatedOT}, \etc.

Previous works tend to exploit various types of consistency among the image group to solve the CoSOD task, including shared features~\cite{CBCS} and common semantic information~\cite{MetricCoSOD}. With the success of unified models in up-stream tasks~\cite{Ren2015FasterRT,OIM}, the latest CoSOD models try to address salient object detection and common object detection in a unified framework~\cite{GCoNet,CoSOD3k,GICD}.
Despite the promising performance achieved by these methods, most of them only focus on learning better consistent feature representations in an individual group~\cite{GICD,GWD,CoADNet,GLNet,ICNet,SRG}, which may make them suffer from the following limitations. First, images from the same group can only act as positive samples of each other. Consensus representations learned from all positive samples might be difficult to be distinguished due to the lack of inter-group separability. Besides, the number of images in a single group is usually insufficient for models to learn robust and unique representations, which can be easily distinguishable from others. Due to the higher complexity of image context in real-world applications, the number of object classes will increase significantly, making the consensuses have a higher risk of getting closer to others and being hard to identify. In this situation, a module designed for bridging the cross-group connection and learning consensuses of distinction is in high demand.

To achieve accurate and fast CoSOD, we propose Memory-aided Contrastive Consensus Learning (\ourmodel{}), which exploits common features within each group and identifies distinctions among different groups, guiding the model to produce co-saliency maps with high integrity. To fulfill the above goal, three key components are proposed in \ourmodel{}. \textbf{First}, we present the Group Consensus Aggregation Module (GCAM) for mining the common feature within the same group by the correlation principle. \textbf{Second}, we introduce the Memory-based Contrastive Module (MCM) to conduct robust contrastive learning with a long-term memory. More concretely, the consensus of each class is saved and updated with momentum in a memory queue to avoid the instability of online contrastive learning. \textbf{Third}, we employ the Adversarial Integrity Learning (AIL) to improve the integrity and quality of predicted maps in an adversarial fashion, where a discriminator identifies whether the masked regions are obtained from predicted or ground-truth maps. Analogous to generative adversarial networks~\cite{GAN}, our model tries to fool the discriminator and produce high-quality and high-integrity maps that can mask complete and intact objects.

Our main contributions can be summarized as follows:
\begin{itemize}

\item We establish a fast yet strong CoSOD baseline with the Transformer, which outperforms most existing methods that are sophisticatedly equipped with many components.

\item We introduce the Group Consensus Aggregation Module (GCAM) to generate the consensus of each group in an effective way. To make the consensus more discriminative to each other, we propose the Memory-based Contrastive Module (MCM) in a metric learning way.

\item Furthermore, the Adversarial Integrity Learning (AIL) is proposed to improve the quality and integrity of predicted co-saliency maps in an adversarial learning manner.

\item We conduct extensive experiments to validate the superiority of our \ourmodel{}. Extensive quantitative and qualitative results show that our \ourmodel{}~can outperform existing CoSOD models by a large margin.

\end{itemize}

\section{Related Work}
\label{sec:Related Work}
\subsection{Salient Object Detection}
Before the deep learning era, handcrafted features played the most critical role in detection~\cite{Cheng2011GlobalCB,SOD_trad1,SOD_trad2} among the traditional SOD methods. When it comes to the early years of deep learning, features are usually extracted from image patches, which will then be used to generate object proposals~\cite{SOD_op1,SOD_op2,SOD_op3}, or super-pixels~\cite{SOD_patch1,SOD_patch2} as processing units. 
As stated in~\cite{PoolNet+}, the network architectures of existing SOD methods can be divided into five categories, \ie, U-shape, side-fusion, multi-branch, single-steam, and multi-stream. So far, U-shape has been the most widely used architecture \cite{UNet}, especially when the fusion between low-level and high-level features is needed.
Supervision on the multi-stage output is also employed at the early stages by aggregating features from different levels of networks in the U-shape architecture to make the output features more robust and stable~\cite{EGNet,GCoNet,GICD}.~\cite{SOD_att1,SOD_att2,SOD_att3} employed the attention mechanism in their models for further improvement. Besides, some external information is also introduced as extra guidance for training, such as boundary~\cite{BASNet}, edge~\cite{EGNet}, and depth~\cite{EGNet}.

\subsection{Co-Salient Object Detection} CoSOD emphasizes on detecting salient objects across groups of images rather than in a single image. Traditional CoSOD methods utilize handcrafted cues (\eg, superpixels~\cite{SLIC}) to explore the correspondence of images. In contrast, the deep learning-based methods learn the consensus feature representation of common objects in an end-to-end manner~\cite{GWD,MetricCoSOD}. Various model architectures are applied to improve the CoSOD performance, including CNN-based~\cite{GCoNet,CoADNet,GICD} and Transformer-based models~\cite{CoSformer}. Though some of the existing methods investigate both intra-group and inter-group cues~\cite{GCoNet}, there is still much room for improvement in the fully coordinated and simultaneous use of intra-group and inter-group information.

\subsection{Integrity Learning for Saliency Maps}
The quality of saliency maps has attracted much attention in recent years to make existing saliency-related tasks closer to real-world applications.
\cite{integrity_1} tries to guide their models to learn integrity via the collaboration between global context and local objects. TSPOANet~\cite{integrity_3} adopts a capsule network to model the part-object relationship to achieve better wholeness and uniformity of segmented salient objects.
In~\cite{BASNet}, a hybrid loss is applied for more focus on improving the boundary of predicted maps. Furthermore,~\cite{integrity} investigates more into the integrity issue in SOD and tries solving this problem with their carefully designed components. In~\cite{GCoNet+}, a confidence enhancement module is proposed to make the predicted maps more binarized.

\section{Methodology}
\label{sec:methodology}

\begin{figure*}[t!]
  \centering
    \begin{overpic}[width=.95\linewidth]{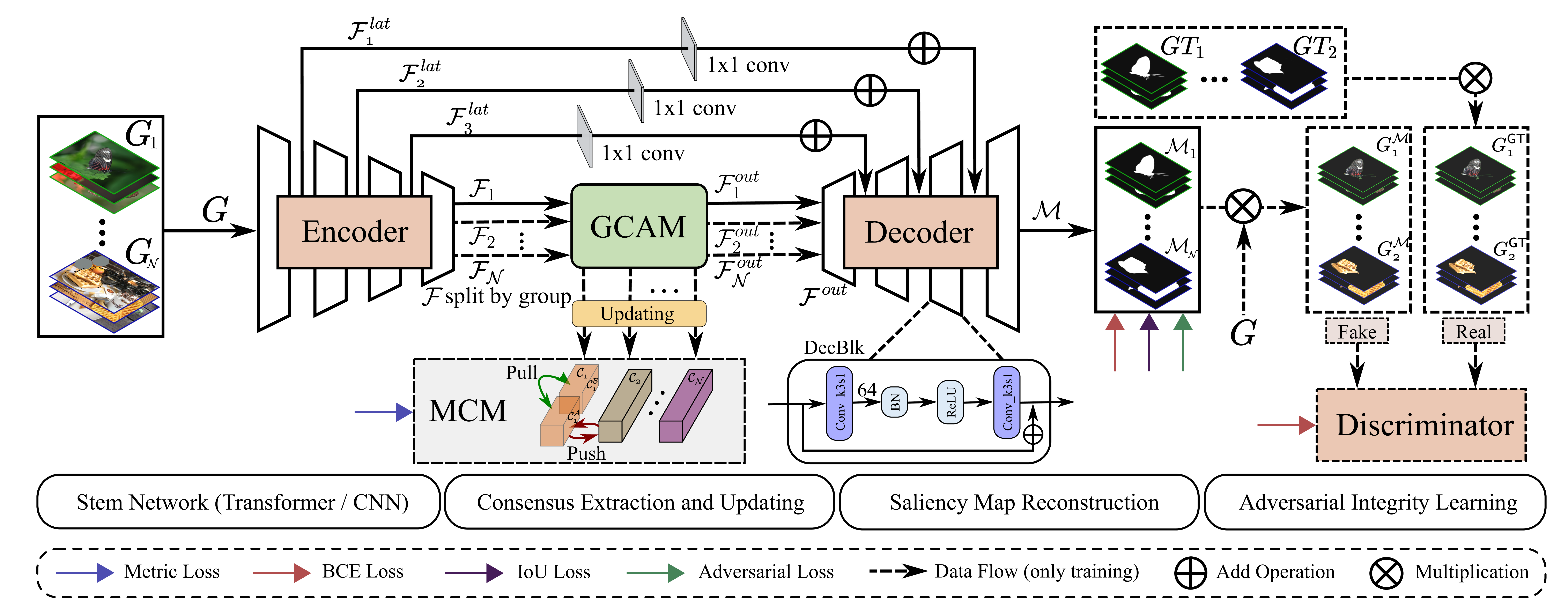}
    \end{overpic}
   \caption{
   \textbf{Overall framework of the proposed Memory-aided Contrastive Consensus Learning (\ourmodel).} Input images are obtained from multiple groups and fed to an encoder. First, we employ the Group Consensus Aggregation Module (GCAM), where the intra-group features of each group can be learned separately. With the consensus learned from each single group, the consensus features are updated in the memory of each class in the queue of Memory-based Contrastive Module (MCM). Then, contrastive learning is conducted to make the consensus more discriminative to each other. Each stage of our encoder and decoder is connected with only a 1x1 convolution layer for feature adding with the least computation. Our decoder is composed of four DecBlk, which is the vanilla residual block. We design our model as simple as possible to make our study more open and solid. Finally, saliency maps of all groups are predicted based on the supervision from the BCE and IoU losses.
   }\label{fig:Arch}
\end{figure*}

In this section, we first introduce the overall architecture of our \ourmodel{}~for the CoSOD task. Then, we sequentially introduce the proposed three key components: Group Consensus Aggregation Module (GCAM), Memory-based Contrastive Module (MCM), and Adversarial Integrity Learning (AIL).

First, GCAM is used to exploit the common features of images in the same group. Second, MCM is applied to make the learned consensus of different groups more robust and discriminative to each other. Finally, we employ AIL to improve the integrity and quality of predicted maps in an adversarial way. Note that MCM and AIL are only used during training and thus can be entirely discarded during inference for a more lightweight model.

\subsection{Overview}
\label{sec:overview}

\figref{fig:Arch} illustrates the basic framework of the proposed \ourmodel{}~including the learning pipeline.
Different from existing CoSOD models that take images from a single group~\cite{CoSOD3k,CoADNet,ICNet,GICD} as input, our model receives images from multiple groups as input, bringing the potential to bridge the intersection between different groups.

First, we take images of $N$ (default as 2) groups as the input \{$G_1$, $G_2$, ..., $G_N$\}. We concatenate all the images as a whole batch $G$, which is then fed to the encoder. With the backbone network (default as the Transformer network PVTv2~\cite{PVTv2}) as our encoder, embedded features are extracted as $\mathcal{F}$, which is then split by their group categories as \{$\mathcal{F}_1, \mathcal{F}_2, ..., \mathcal{F}_N$\}, where $\mathcal{F}_N=\{F_{N,s}\}_{s=1}^S \in \mathbb{R}^{S \times C \times H \times W}$, $C$ denotes the channel number, $H \times W$ means the spatial size, and $N$ is the group size. Meanwhile, the intermediate features \{$\mathcal{F}_1^{lat}, \mathcal{F}_2^{lat}, \mathcal{F}_3^{lat}$\} at different stages of our encoder are saved and fed to their corresponding stages of the decoder by a 1x1 convolutional layer.

Then, \{$\mathcal{F}_1, \mathcal{F}_2, ..., \mathcal{F}_N$\} are sequentially fed to GCAM to obtain the consensus of each group. With the consensus of groups \{$\mathcal{F}_1^{out}, \mathcal{F}_2^{out}, ..., \mathcal{F}_N^{out}$\}, the memory of the corresponding classes is updated in the queue with momentum, supervised by a metric learning loss used in~\cite{GCoNet+}. 

Furthermore, the consensus of all groups is concatenated as $\mathcal{F}^{out}$ and fed to the decoder, which consists of four stacked standard residual blocks and combines the early features from lateral connections. Co-saliency maps $\mathcal{M}$ are generated at the end of the decoder. The prediction of the decoder $M$ is supervised by the Binary Cross Entropy (BCE) loss and the Intersection over Union (IoU) loss, which provide pixel-level and region-level supervision, respectively.

Finally, the predicted co-saliency maps $\mathcal{M}$ are facilitated together with the source images $G$ and the ground-truth maps $GT$. The pixel-wise multiplication between $G$ and $M$ leads to $G^\mathcal{M}$, and we obtain $G^{GT}$ in a similar way. $G^\mathcal{M}$ and $G^{GT}$ are then fed to an independent discriminator, identifying whether the masked images are generated by the ground-truth maps $G^{GT}$ or the source images $G$, which include intact and complete objects. Accordingly, the adversarial loss from the discriminator is applied to the whole generator, and the BCE loss is given to the discriminator.

\subsection{Group Consensus Aggregation Module}
\label{sec:GCAM}

\begin{figure}[t!]
  \centering
    \begin{overpic}[width=.75\linewidth]{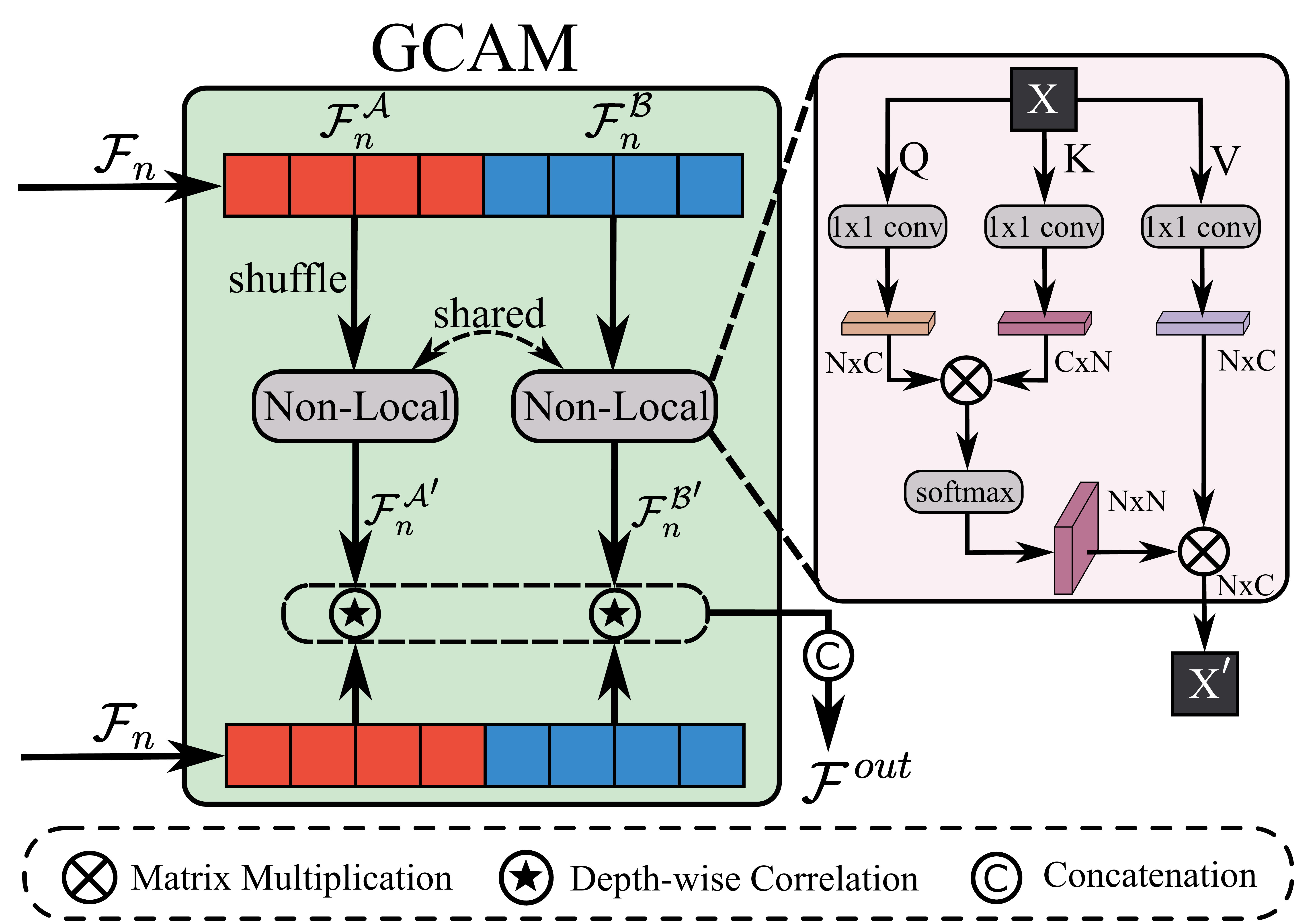}
    \end{overpic}
   \caption{
   \textbf{Group Consensus Aggregation Module.} The feature of the encoder is fed to the GCAM and handled group by group. The original features of one group are evenly split and shuffled before being fed to the non-local block. The depth-wise correlation bridges the semantic interaction between the consensus and original features.
   }\label{fig:GCAM}
\end{figure}

In the wild, objects of the same category tend to share similar appearance, which has been exploited in many related tasks, such as video tracking~\cite{wang2019learning} and semantic segmentation~\cite{Zhang2019CoOccurrentFI}, where the correspondence between common objects is used as prior information. Here we also apply this mechanism to CoSOD. Similar to~\cite{GCoNet}, we employ the non-local block~\cite{non_local} to extract the affinity feature. As shown in \figref{fig:GCAM}, we first split the output feature of the encoder $\mathcal{F}_n$ into \{$\mathcal{F}_n^A, \mathcal{F}_n^B$\}, which are then shuffled and fed into the non-local block. Subsequently, in the non-local block, we compute the affinity map of the feature and conduct matrix multiplication between the affinity map and the value feature (\ie, `V' in the non-local block) to obtain the consensus feature \{${\mathcal{F}_n^A}^\prime, {\mathcal{F}_n^B}^\prime$\}. Finally, we perform depth-wise correlation to fuse the original feature with the consensus feature, and concatenate them to form the final consensus representation $\mathcal{F}_{out}$.

\subsection{Memory-based Contrastive Module}
\label{sec:MCM}

Metric learning is a widely-used technique that contributes to distinguishing features of different clusters and works in many tasks, including CoSOD~\cite{MetricCoSOD,MetricCoSOD1,GCoNet+}. However, CoSOD datasets only contain a limited number of images (tens of images) of limited groups (less than 300 groups). In such cases, naive metric learning cannot work very well due to the small number of samples insufficient for distance measurement.

To overcome this issue, some contrastive learning methods introduce the memory queue to achieve more robust contrastive learning with a long-term memory, such as MoCo~\cite{MOCO}, OIM~\cite{OIM}, \etc{}. Inspired by these works, we propose the MCM, which saves the consensus features of each class into memory blocks and updates the corresponding blocks with momentum in every batch. To be more specific, as shown in \figref{fig:Arch}, the consensuses of all groups \{$\mathcal{F}_1^{out}, \mathcal{F}_2^{out}, ..., \mathcal{F}_N^{out}$\} are saved or updated in their own memory blocks as \{$\mathcal{C}_1, \mathcal{C}_2^, ..., \mathcal{C}_N$\}. The memory update is as follows:
\begin{equation}
  \mathcal{C}^t_1 = \beta * \mathcal{C}^{t-1}_1 + (1 - \beta) * \mathcal{F}_1^{out},
\end{equation}
where $\beta$ denotes the momentum factor and is set to 0.1 by default. When $\beta$ is set to 0, the MCM belongs to fully online metric learning.

As demonstrated in the MCM, each memory block splits itself into two parts, $\mathcal{C}_1^{A}$ and $\mathcal{C}_1^{B}$. In this case, $\mathcal{C}_1^{B}$ is viewed as the positive samples of $\mathcal{C}_1^{A}$, and the whole $\mathcal{C}_2$ is considered as the negative samples of $\mathcal{C}_1^{A}$~\cite{GCoNet+}. Then, the loss of MCM can be computed by the GST loss~\cite{GCoNet+} as follows:
\begin{equation}
  L_\mathrm{Triplet}(C_1, C_2) = ||\mathcal{C}_1^A - \mathcal{C}_1^B||_2 - ||\mathcal{C}_1^A - \mathcal{C}_2^B||_2 + \alpha 
\end{equation}
\begin{equation}
  L_\mathrm{MCM} = \frac{1}{N^2}\sum_{i=1}^{N}\sum_{j=1}^{N}{L_\mathrm{Triplet}(C_i, C_j)},
\end{equation}
where $\alpha$ denotes the margin used in the triplet loss~\cite{TripletLoss}, which is set to 0.1. $|| \cdot ||_2$ means the $l_2$ norm of the input.

\subsection{Adversarial Integrity Learning}
\label{sec:AIL}

Although some latest works have investigated the integrity of SOD~\cite{integrity}, they try to solve this problem by designing sophisticated model architectures and critical components to make predicted maps of higher integrity. These attempts may lead to maps with better quality, but the motivation of their design is not intuitive to the integrity problem.

\begin{figure}[t!]
  \centering
    \begin{overpic}[width=.95\linewidth]{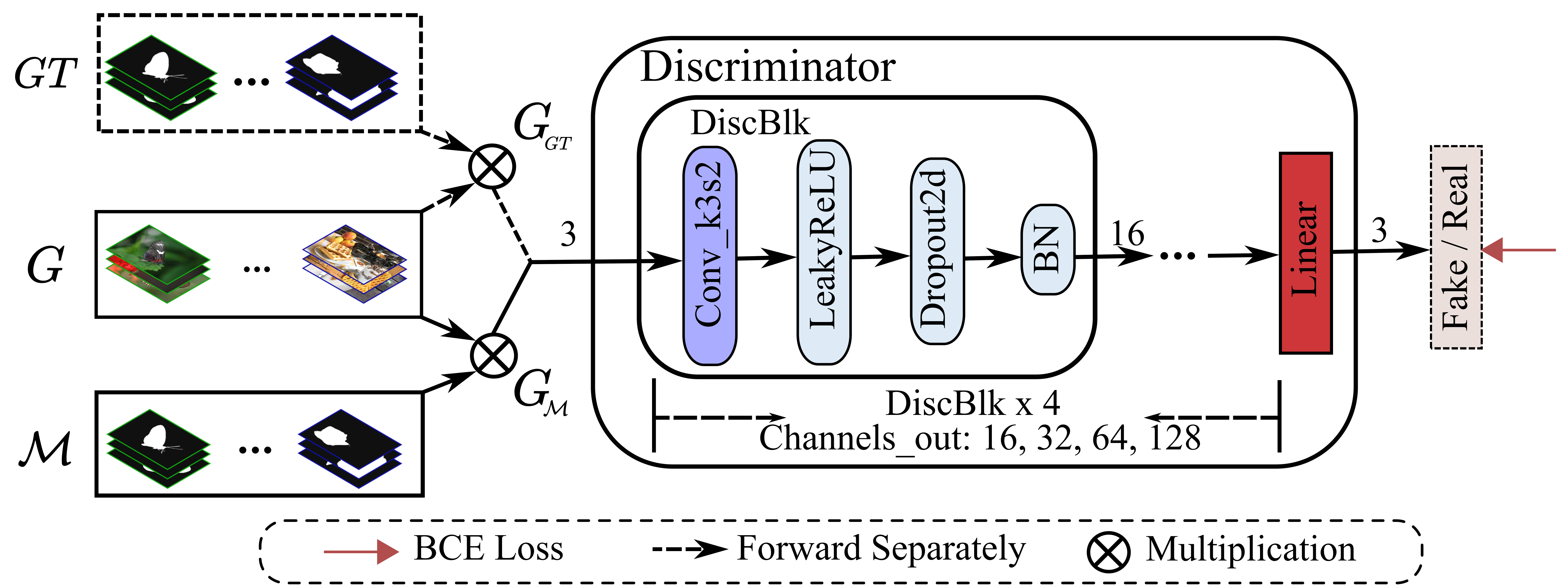}
    \end{overpic}
   \caption{
   \textbf{Discriminator used in our AIL.} The discriminator has four discrimination blocks (DiscBlk) stacked sequentially, with 16, 32, 64, and 128 as the number of their output channels, respectively. Note that there is no batch normalization layer in the first DiscBlk in our implementation.
   }\label{fig:Disc}
\end{figure}

To explicitly solve this problem, we propose the Adversarial Integrity Learning (AIL) in our framework. Three data sources exist in AIL, \ie{}, the source images, the ground-truth maps, and the predicted maps in the current batch. During training, we perform pixel-wise multiplications on two pairs, \ie{}, (source images, ground-truth maps) and (source images, predicted maps), as shown in \figref{fig:Disc}, to obtain $G_{GT}$ and $G_M$, respectively. Then, we employ a discriminator to identify whether the regions of source images masked by these two maps are real or fake, as shown in \figref{fig:Arch}. Obviously, the regions masked by the ground-truth maps are complete and intact objects with 100\% integrity. During training, the loss from the discriminator guides the generator to produce maps that can localize objects with higher accuracy and integrity. The ablation results are shown in \figref{fig:abla_ail}.

\subsection{Objective Function}
As shown in \figref{fig:Arch}, the objective function $L_{sal}$ of the main network (generator) is a weighted combination of the low-level losses (\ie, BCE and IoU losses) and the high-level losses (metric and adversarial losses). And the discriminator involves the BCE loss. The details of $L_{MCM}$ can be found in the `Methodology' section above.
The BCE, IoU, and adversarial losses are as follows:
\begin{equation}
  {L}_\mathrm{BCE} = - \sum{[Y log(\hat{Y}), (1 - Y)log(1 - \hat{Y})]},
  \label{eqn:loss_BCE}
\end{equation}
\begin{equation}
  {L}_\mathrm{IoU} = 1 - \frac{1}{N}\sum{\frac{Y \cap \hat{Y}}{Y \cup \hat{Y}}},
  \label{eqn:loss_IoU}
\end{equation}
where $Y$ is the ground-truth map, $\hat{Y}$ is the predicted map.
\begin{equation}
  \hat{T} = discriminator(\hat{Y} \cdot G),
  \label{eqn:loss_adv_g_0}
\end{equation}
\begin{equation}
  {L}_\mathrm{adv} = - \sum{[T log(\hat{T}), (1 - T)log(1 - \hat{T})]},
  \label{eqn:loss_adv_g_1}
\end{equation}
where $\hat{Y}$ is the predicted map, $G$ denotes the source images, $\cdot$ denotes the pixel-wise multiplication, $\hat{T}$ and $T$ denote the prediction of discriminator on whether $\hat{Y}$ and $Y$ is the ground-truth map, respectively. 

Therefore, our final objective function is:
\begin{equation}
  {L_\mathrm{sal}} = \lambda_1 {L}_\mathrm{BCE} + \lambda_2 {L}_\mathrm{IoU} + \lambda_3 {L}_\mathrm{MCM} + \lambda_4 {L}_\mathrm{adv},
  \label{eqn:loss_final}
\end{equation}
\begin{equation}
  {L_\mathrm{disc}} = \lambda_5 {L}_\mathrm{BCE},
  \label{eqn:loss_disc}
\end{equation}
where $\lambda_1, \lambda_2, \lambda_3, \lambda_4$, and $\lambda_5$ are respectively set to 30, 0.5, 3, 10, and 3 to keep all the losses at a reasonable scale at the beginning of training to benefit the optimization.

\section{Experiments}

\subsection{Datasets}
\label{sec:exp_datasets}
\textbf{Training Sets.} We follow \cite{CoADNet} to use DUTS\_class~\cite{GICD} and COCO-SEG~\cite{COCO_SEG} as our training sets. The whole DUTS\_class is divided into 291 groups, which contain 8,250 images in total. COCO-SEG contains 200k images of 78 groups and corresponding binary maps.

\begin{table*}[t!]
\begin{center}
\footnotesize
\renewcommand{\arraystretch}{1.0}
\renewcommand{\tabcolsep}{1.14mm}
\begin{tabular}{r||c|cccc|cccc|cccc}
\hline
&  & \multicolumn{4}{c|}{CoCA~\cite{GICD}} & \multicolumn{4}{c|}{CoSOD3k~\cite{CoSOD3k}} & \multicolumn{4}{c}{CoSal2015~\cite{CoSal2015}} \\
\multirow{-2}{*}{Method} & \multirow{-2}{*}{Pub.} & $E_{\xi}^\mathrm{max} \uparrow$ & $S_\alpha \uparrow$ & $F_\beta^\mathrm{max} \uparrow$ & $\epsilon \downarrow$ & $E_{\xi}^\mathrm{max} \uparrow$ & $S_\alpha \uparrow$ & $F_\beta^\mathrm{max} \uparrow$ & $\epsilon \downarrow$ & $E_{\xi}^\mathrm{max} \uparrow$ & $S_\alpha \uparrow$ & $F_\beta^\mathrm{max} \uparrow$ & $\epsilon \downarrow$ \\
\hline
CBCS~\shortcite{CBCS} & TIP & 0.641 & 0.523 & 0.313 & 0.180 & 0.637 & 0.528 & 0.466 & 0.228 & 0.656 & 0.544 & 0.532 & 0.233 \\
GWD~\shortcite{GWD} & IJCAI & 0.701 & 0.602 & 0.408 & 0.166 & 0.777 & 0.716 & 0.649 & 0.147 & 0.802 & 0.744 & 0.706 & 0.148 \\
RCAN~\shortcite{RCAN} & IJCAI & 0.702 & 0.616 & 0.422 & 0.160 & 0.808 & 0.744 & 0.688 & 0.130 & 0.842 & 0.779 & 0.764 & 0.126 \\
GCAGC~\shortcite{GCAGC} & CVPR & 0.754 & 0.669 & 0.523 & 0.111 & 0.816 & 0.785 & 0.740 & 0.100 & 0.866 & 0.817 & 0.813 & 0.085 \\
GICD~\shortcite{GICD} & ECCV & 0.715 & 0.658 & 0.513 & 0.126 & 0.848 & 0.797 & 0.770 & 0.079 & 0.887 & 0.844 & 0.844 & 0.071 \\
ICNet~\shortcite{ICNet} & NeurIPS & 0.698 & 0.651 & 0.506 & 0.148 & 0.832 & 0.780 & 0.743 & 0.097 & 0.900 & 0.856 & 0.855 & 0.058 \\
CoADNet~\shortcite{CoADNet} & NeurIPS & - & - & - & - & 0.878 & 0.824 & 0.791 & 0.076 & 0.914 & 0.861 & 0.858 & 0.064 \\
DeepACG~\shortcite{DeepACG} & CVPR & 0.771 & 0.688 & 0.552 & 0.102 & 0.838 & 0.792 & 0.756 & 0.089 & 0.892 & 0.854 & 0.842 & 0.064 \\
GCoNet~\shortcite{GCoNet} & CVPR & 0.760 & 0.673 & 0.544 & 0.105 & 0.860 & 0.802 & 0.777 & 0.071 & 0.887 & 0.845 & 0.847 & 0.068 \\
CoEGNet~\shortcite{CoSOD3k} & TPAMI & 0.717 & 0.612 & 0.493 & 0.106 & 0.837 & 0.778 & 0.758 & 0.084 & 0.884 & 0.838 & 0.836 & 0.078 \\
CADC~\shortcite{CADC} & ICCV & 0.744 & 0.681 & 0.548 & 0.132 & 0.840 & 0.801 & 0.859 & 0.096 & 0.906 & 0.866 & 0.862 & 0.064 \\
DCFM$^*$~\shortcite{DCFM} & CVPR & 0.783 & 0.710 & \textbf{0.598} & \textbf{0.085} & 0.874 & 0.810 & 0.805 & 0.067 & 0.892 & 0.838 & 0.856 & 0.067 \\
UFO~\shortcite{arxiv_UFO} & arXiv & 0.782 & 0.697 & 0.571 & 0.095 & 0.874 & 0.819 & 0.797 & 0.073 & 0.906 & 0.860 & 0.865 & 0.064 \\
\hline
\rowcolor{mygray} \textbf{Ours} & Sub. & \textbf{0.796} & \textbf{0.714} & 0.590 & 0.103 & \textbf{0.903} & \textbf{0.858} & \textbf{0.837} & \textbf{0.061} & \textbf{0.927} & \textbf{0.890} & \textbf{0.891} & \textbf{0.051} \\
\hline
\end{tabular}
\caption{\textbf{Quantitative comparisons between our \ourmodel{}~and other methods.} ``$\uparrow$'' (``$\downarrow$'') means that the higher (lower) is better. $^*$ denotes the state-of-the-art method. UFO~\cite{arxiv_UFO} is still an arXiv paper and does not show much improvement compared with DCFM~\cite{DCFM}, so we set DCFM as the previous SoTA.}
\label{tab:sota}
\end{center} 
\end{table*}

\textbf{Test Sets.} 
For a comprehensive evaluation of our \ourmodel{}, we test it on three widely used CoSOD datasets, \ie, CoCA~\cite{GICD}, CoSOD3k~\cite{CoSOD3k}, and CoSal2015~\cite{CoSal2015}. Among the three datasets, CoCA is the most challenging one. 
It is of much higher diversity and complexity in terms of background, occlusion, illumination, surrounding objects, \etc. Following the latest benchmark~\cite{CoSOD3k}, we do not evaluate on iCoseg~\cite{iCoseg} and MSRC~\cite{MSRC}, since only one salient object is given in most images there. It is more convincing to evaluate CoSOD methods on images with more salient objects, which is closer to real-life applications.

\subsection{Evaluation Protocol}
Following GCoNet~\cite{GCoNet}, we employ the S-measure~\cite{Smeasure}, maximum F-measure~\cite{Fmeasure}, maximum E-measure~\cite{Emeasure}, and mean absolute error (MAE) to evaluate the performance in our experiments.

\subsection{Implementation Details}

We select samples from DUTS\_class~\cite{GICD} and COCO-SEG~\cite{COCO_SEG} alternatively, and set the batch size as follows:
\begin{equation}
  batch size = min(\#group 1, ..., \#group N, 48),
  \label{eqn:batch_size}
\end{equation}
where $\#$ means the image number in the corresponding group. The images are resized to 256x256 for training and inference. The output maps are resized to the original size for evaluation. We apply three data augmentation strategies, \ie, horizontal flip, color enhancement, and rotation. Our \ourmodel{}~is trained for 250 epochs with the AdamW optimizer~\cite{AdamW}. The initial learning rate is 1e-4 and divided by 10 at the last 20th epoch. The whole training process takes $\sim$3.5 hours and consumes only $\sim$7.5GB GPU memory. All the experiments are implemented based on the PyTorch library~\cite{PyTorch} with a single NVIDIA RTX3090 GPU.

\subsection{Ablation Study}
\label{sec:Ablation}
\begin{figure}[t!]
	\flushright
    \begin{overpic}[width=.95\linewidth]{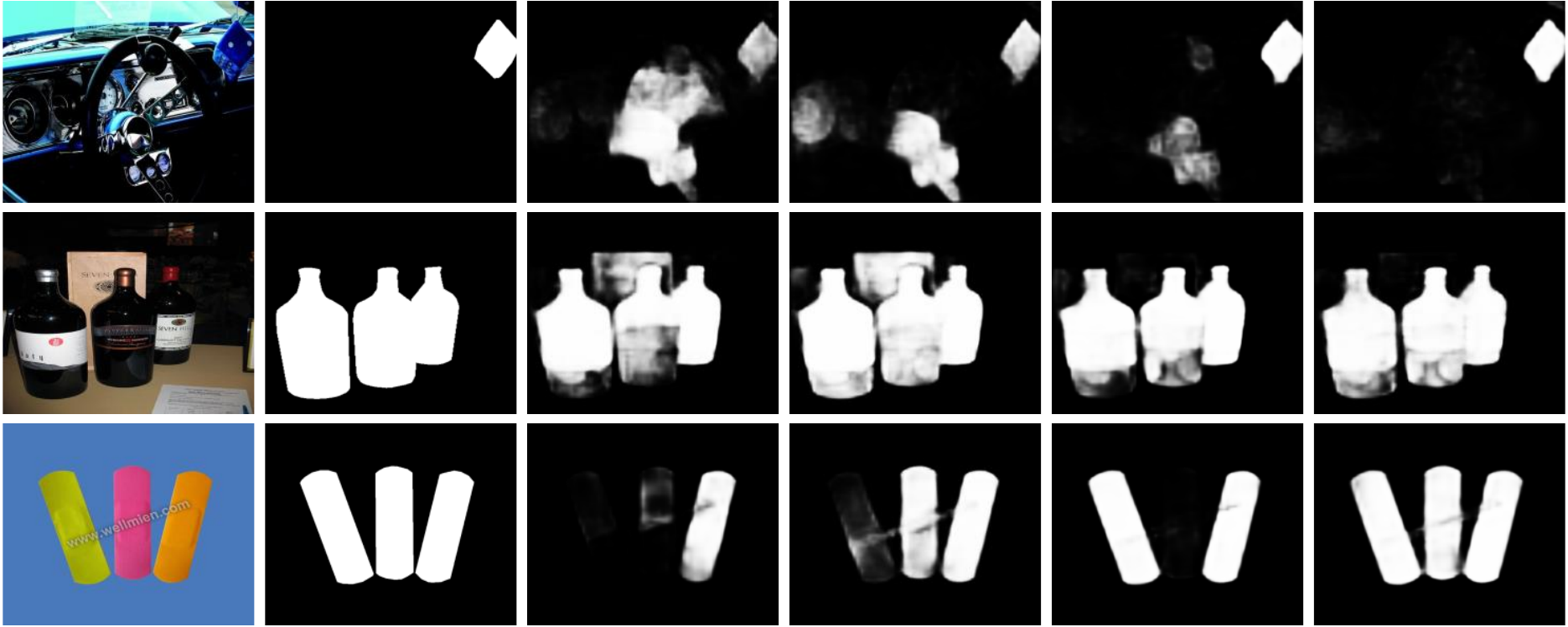}
        
        \put(-4,29){\rotatebox{90}{Dice}}
        \put(-4,15){\rotatebox{90}{Bottle}}
        \put(-4,2){\rotatebox{90}{Bind}}
        \put(6,-4.5){{(a)}}
        \put(22.5,-4.5){{(b)}}
        \put(39,-4.5){{(c)}}
        \put(56,-4.5){{(d)}}
        \put(73,-4.5){{(e)}}
        \put(90,-4.5){{(f)}}
    \end{overpic}
	\vspace{10pt}
	\caption{\textbf{Qualitative ablation studies of different modules and their combinations in our \ourmodel.} (a) Source image; (b) Ground truth; (c) w/ GCAM; (d) w/ GCAM+MCM; (e) w/ GCAM+AIL; (f) w/ GCAM+MCM+AIL, \ie, the full version of our model.}
	\label{fig:qualitive_ablation}
\end{figure}
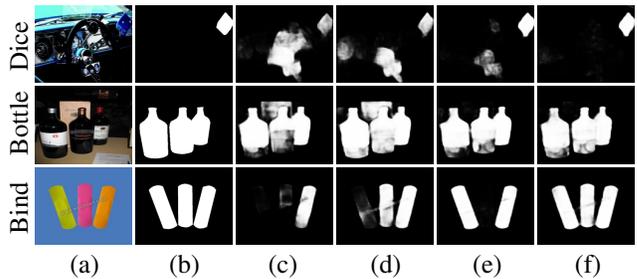

We conduct the ablation study to validate the effectiveness of each component (\ie, GCAM, MCM, and AIL) employed in our \ourmodel{}.
The qualitative results regarding each module and the combination are shown in \figref{fig:qualitive_ablation}.

\begin{table*}[t!]
\begin{center}
\footnotesize
\renewcommand{\arraystretch}{1.0}
\setlength\tabcolsep{4.5pt}
\begin{tabular}{ccc||cccc|cccc|cccc}
\hline
\multicolumn{3}{c||}{Modules}  & \multicolumn{4}{c|}{CoCA~\cite{GICD}} & \multicolumn{4}{c|}{CoSOD3k~\cite{CoSOD3k}} & \multicolumn{4}{c}{CoSal2015~\cite{CoSal2015}} \\
\hspace{1.25mm}GCAM\hspace{1.25mm} & MCM & AIL & $E_{\xi}^\mathrm{ max} \uparrow$ & $S_\alpha \uparrow$ & $F_\beta^\mathrm{ max} \uparrow$ & $\epsilon \downarrow$ & $E_{\xi}^\mathrm{ max} \uparrow$ & $S_\alpha \uparrow$ & $F_\beta^\mathrm{ max} \uparrow$ & $\epsilon \downarrow$ & $E_{\xi}^\mathrm{ max} \uparrow$ & $S_\alpha \uparrow$ & $F_\beta^\mathrm{ max} \uparrow$ & $\epsilon \downarrow$ \\
\hline
 &  &  & 0.756 & 0.683 & 0.553 & 0.118 & 0.880 & 0.828 & 0.798 & 0.075 & 0.905 & 0.866 & 0.861 & 0.062 \\
\checkmark &  &  & 0.779 & 0.709 & 0.577 & 0.103 & 0.894 & 0.851 & 0.824 & 0.061 & 0.921 & 0.884 & 0.882 & 0.053 \\
\checkmark & \checkmark &  & 0.788 & 0.711 & 0.585 & 0.100 & 0.898 & 0.853 & 0.828 & 0.060 & 0.925 & 0.889 & 0.886 & \textbf{0.050} \\

\checkmark &  & \checkmark & 0.789 & 0.714 & 0.585 & \textbf{0.097} & 0.900 & 0.855 & 0.831 & 0.060 & 0.924 & 0.888 & 0.887 & 0.053 \\
\hline
\rowcolor{mygray}
\checkmark & \checkmark & \checkmark & \textbf{0.796} & \textbf{0.714} & \textbf{0.590} & 0.103 & \textbf{0.903} & \textbf{0.858} & \textbf{0.837} & \textbf{0.061} & \textbf{0.927} & \textbf{0.890} & \textbf{0.891} & 0.051 \\
\hline
\end{tabular}
\caption{\textbf{Quantitative ablation studies of the proposed components in our \ourmodel{}.} The components include the GCAM, MCM, AIL, and their combinations.}
\label{tab:ablation_modules}
\end{center}
\end{table*}

\begin{figure*}[t!]
	\centering
    \begin{overpic}[width=.96\textwidth]{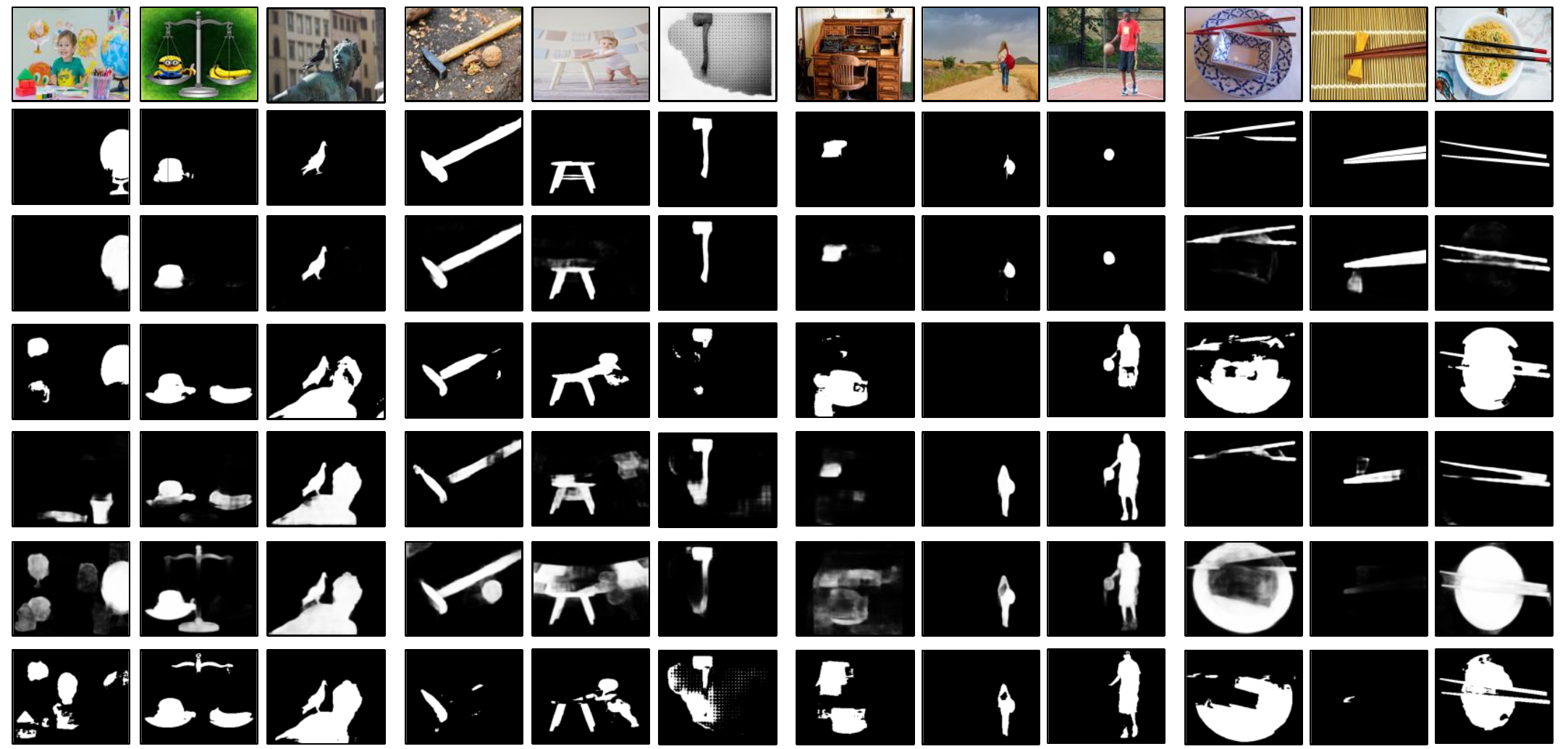}
        \put(-1.5,42){\rotatebox{90}{Input}}
        \put(-1.5,36){\rotatebox{90}{GT}}
        \put(-1.5,29){\rotatebox{90}{\textbf{Ours}}}
        \put(-1.5,21){\rotatebox{90}{DCFM}}
        \put(-1.5,15){\rotatebox{90}{UFO}}
        \put(-1.5,7){\rotatebox{90}{CADC}}
        \put(-1.5,-0.5){\rotatebox{90}{GCoNet}}
        \put(6,-2){{Salient Distractors}}
        \put(29,-2){{Complex Structure}}
        \put(57,-2){{Tiny Objects}}
        \put(77,-2){{Extremely Difficult Cases}}
    \end{overpic}
	\vspace{10pt}
	\caption{\textbf{Qualitative comparisons of our \ourmodel{} and other methods.} `GT' denotes the ground truth. We select the results of hard cases due to various reasons. The `Extremely Difficult Cases' means the chopstick group on the test set of CoCA~\cite{GICD}, as chopsticks are thin, tall, and hard to detect. This could be the most difficult case among all groups on the existing test sets.}
	\label{fig:qual}
\end{figure*}
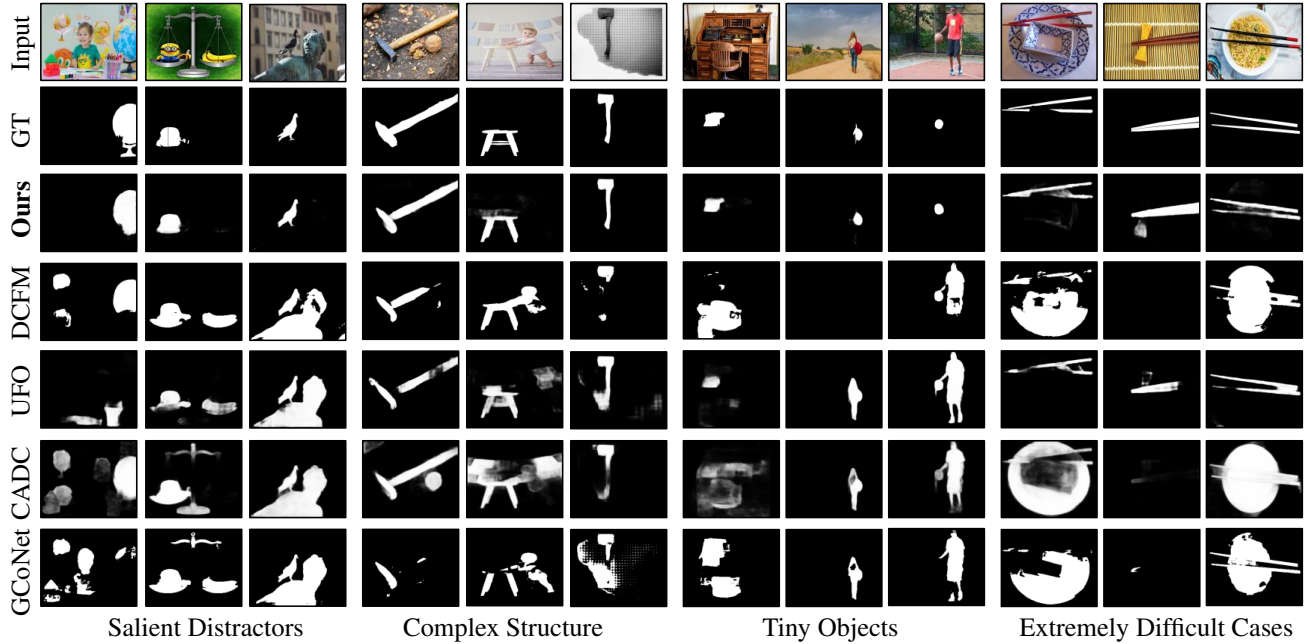

\textbf{Baseline.} We establish a solid CoSOD network as the baseline. To keep pace with the latest Transformer network, we also build our model with both Transformer and convolutional neural networks as the backbone. Following GCoNet~\cite{GCoNet}, we feed images of multiple classes and their ground-truth as the input to train our \ourmodel{}. Compared with previous CoSOD models, our baseline network achieves promising performance with a simpler architecture and much higher speed. To be consistent with the widely used Transformer network~\cite{ViT,PVTv1,PVTv2}, in contrast with the previous CoSOD models~\cite{GICD,GCoNet,CADC,DCFM}, we make our model shallower and build it with only four stages in its encoder and decoder. To achieve a pure and fast baseline network, we firstly substitute all the complex blocks in each lateral connection used in~\cite{GICD,GCoNet,CADC,DCFM} with a single 1x1 convolution layer~\cite{FPN} as the vanilla Feature Pyramid Network (FPN)~\cite{FPN} does. Secondly, we set only a single residual block as the decoding block, where the output is added with the features from the lateral connection. Finally, instead of the multi-stage supervision on all stages of the decoder~\cite{GICD,GCoNet,DCFM}, we set the pixel-level supervision on only the final output with a weighted sum of BCE and IoU losses to guide the model locally and globally, respectively. Our baseline can beat most of the existing CoSOD methods, and thus can be referred to as a strong baseline for others in the future.

\textbf{GCAM.} As the performance evaluated in \tabref{tab:ablation_modules}, our GCAM brings much improvement not only on CoCA and CoSOD3k that focus more on complex context with multiple objects, but also on CoSal2015 that is a relatively simple dataset but needs more attention on the precise SOD in a simple environment.

\textbf{MCM.} In \tabref{tab:ablation_modules}, MCM shows its consistent improvement in all metrics on all datasets. As shown in \figref{fig:qualitive_ablation}, MCM helps our model make more accurate predictions than those models without it.

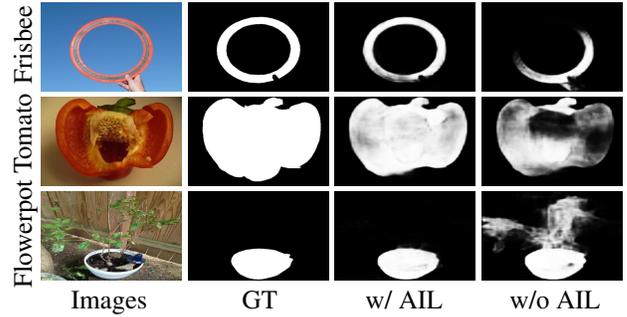
\begin{figure}[t!]
  \flushright
    \begin{overpic}[width=.95\linewidth]{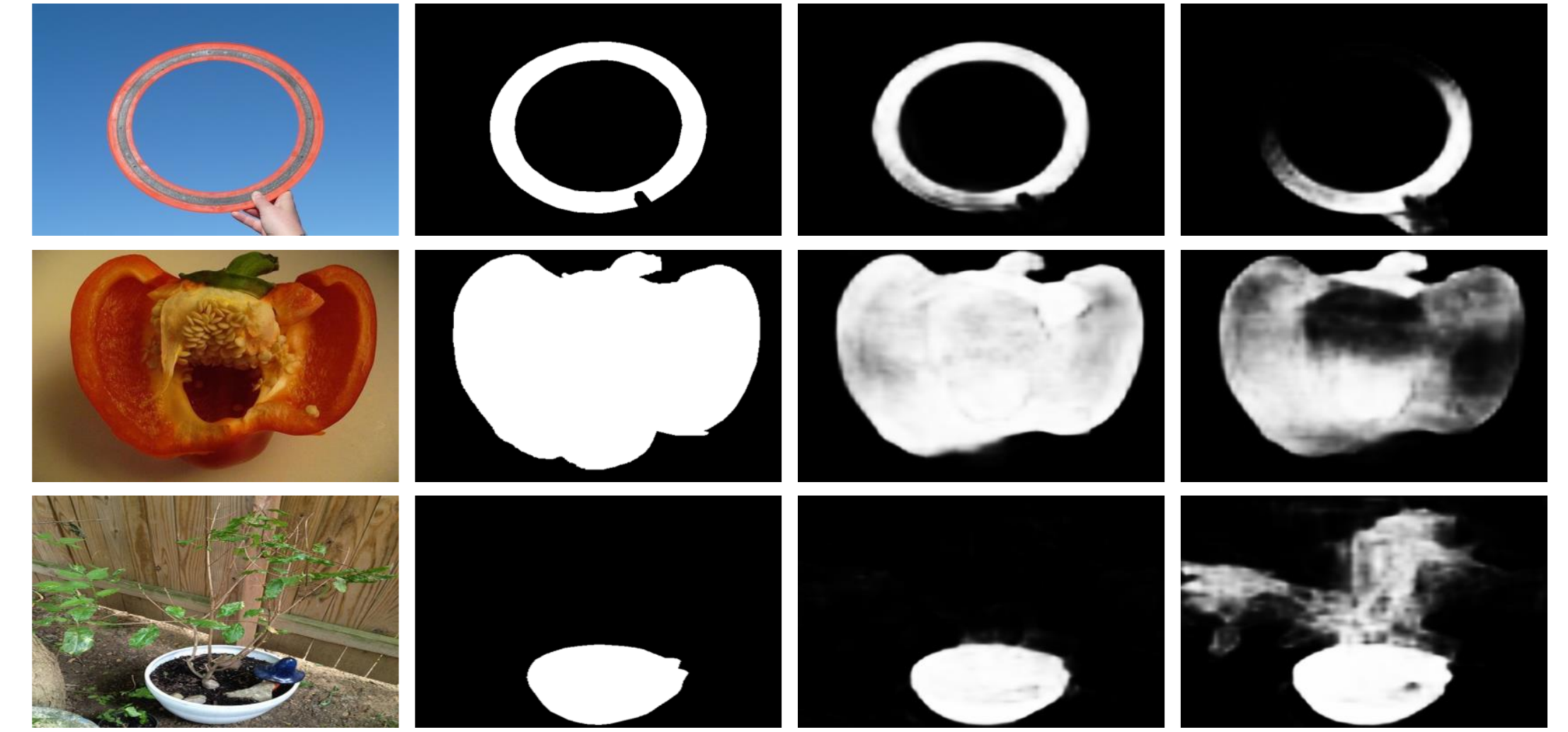}
        \put(-2.5,33){\rotatebox{90}{Frisbee}}
        \put(-2.5,18){\rotatebox{90}{Tomato}}
        \put(-2.5,-1){\rotatebox{90}{Flowerpot}}
        \put(7,-4.5){{Images}}
        \put(35.5,-4.5){{GT}}
        \put(56,-4.5){{w/ AIL}}
        \put(80,-4.5){{w/o AIL}}
    \end{overpic}
    \vspace{10pt}
    \caption{
    \textbf{Qualitative ablation studies of our AIL.} We conduct the qualitative comparison between the baseline model with (w/) and without (w/o) the proposed AIL.
    }\label{fig:abla_ail}
\end{figure}

\textbf{AIL.} 
AIL guides our model to learn the integrity of predicted regions, and the produced co-saliency maps tend to be more robust and contain one or more complete and intact objects. As shown in \figref{fig:abla_ail}, the improvement brought by AIL can be seen from three perspectives. 1) On the object level, AIL increases the completeness of predicted maps of slender objects, which is usually hard to be fully detected. 2) Inside the object, AIL helps fill the unconfident regions which break the structural integrity of the detected objects. 3) Outside the object, AIL suppresses the distractors, with which the regions include just not a complete object. 

In summary, 1) GCAM splits features into two parts, accelerating the affinity generation. 2) MCM is initially motivated by OIM~\cite{OIM}. Differently, MCM saves features by classes instead of identities; OIM uses the final normalized features to update the queue, while we use the consensus generated by GCAM, which aligns well with CoSOD. 3) The adversarial learning strategy in AIL can also be found in domain adaptation~\cite{adv_da}, but we are the first to accommodate it to SOD and to use the segmented regions for discrimination. We also apply AIL to other CoSOD models to validate its high generalizability, as shown in~\tabref{tab:ablation_ail}.

\begin{table}[t!]
\begin{center}
\footnotesize
\begin{tabular}{cc|cccc}
\hline
\multicolumn{1}{c}{CoSOD Models}  & \multicolumn{1}{c|}{AIL}  & \multicolumn{4}{c}{CoCA~\cite{GICD}} \\
\hline
GCoNet~\shortcite{GCoNet} &  & 0.760 & 0.673 & 0.544 & 0.105 \\
GCoNet~\shortcite{GCoNet} & \checkmark & 0.777 & 0.681 & 0.549 & 0.106 \\
GICD~\shortcite{GICD} &  & 0.715 & 0.658 & 0.513 & 0.126 \\
GICD~\shortcite{GICD} & \checkmark & 0.718 & 0.675 & 0.524 & 0.127 \\
\hline
\end{tabular}
\caption{\textbf{Quantitative ablation studies of the proposed AIL on different models.} We apply the proposed AIL to other CoSOD models~\cite{GCoNet,GICD} and conduct the evaluation on CoCA~\cite{GICD}.}
\label{tab:ablation_ail}
\end{center}
\end{table}

\subsection{Comparison to State-of-the-Art Methods}
To make a comprehensive comparison, we compare our \ourmodel{} with one traditional classical algorithm CBCS~\cite{CBCS} and 12 update-to-date deep learning based CoSOD models (see \tabref{tab:sota} for all methods used for comparison). 
Since CoSOD methods have gained much improvement in the past few years and obtained much better performance compared with single-SOD methods, we do not list the single-SOD ones, following previous works~\cite{CoSOD3k,GCoNet,GICD,DCFM}. The detailed leaderboard of previous methods can be found in~\cite{CoSOD3k}.

\textbf{Quantitative Results.} \tabref{tab:sota} shows the quantitative results of our \ourmodel{} and previous competitive methods. Given the above results, we can see that our \ourmodel{} outperforms all the existing methods, especially on the CoSOD3k~\cite{CoSOD3k} and CoSal2015~\cite{CoSal2015} datasets, where the ability to detect salient objects is in a higher priority than finding objects with the common class.

\textbf{Qualitative Results.} \figref{fig:qual} shows the co-saliency maps predicted by different methods for a clear qualitative comparison, where we provide four different types of complex samples from CoCA~\cite{GICD} and CoSOD3k~\cite{CoSOD3k}. Compared with existing models, our~\ourmodel{} shows a stronger ability to eliminate distractors, detect tiny targets, and handle the objects that blend into complex scenes. The extremely difficult cases in which other up-to-date methods fail most of the time further demonstrate the more robust performance of our \ourmodel{}.

%
%

\section{Conclusion}
\label{sec:conclusion}
In this paper, we investigate a novel memory-aided contrastive consensus learning framework (\ie, \ourmodel) for CoSOD. As experiments show, the memory-based contrastive learning with group consensus works effectively to enhance the representation capability of the obtained group features. Besides, the adversarial integrity learning strategy does benefit the saliency model, with the potential to improve the integrity and quality of saliency maps for a variety of SOD and CoSOD models in a general way.

\clearpage

\bibliography{aaai23}

\end{document}